\documentclass{article}
\usepackage{spconf,amsmath,graphicx,amsfonts,booktabs,url}

\usepackage{enumitem}
\setlist{nosep, leftmargin=14pt}

\title{Interpretable Context-Aware Multi-Label Retinal Classification with Dual-Hub Heterogeneous Graph Neural Network}
\name{Proshanto Chanda$^\star$, Munib Mesinovic$^\star$, Tingting Zhu$^\star$, Abhirup Banerjee$^\star$}
\address{$^\star$Institute of Biomedical Engineering, Department of Engineering Science, University of Oxford, UK}
\begin{document}
%
\maketitle
\begin{abstract}
Multi-label retinal screening requires capturing co-occurring pathologies and clinical context, yet standard multimodal fusion remains opaque. We propose a dual-hub heterogeneous graph neural network (GNN) that couples spatial fundus features and structured clinical covariates through cross-modal message passing while providing inherent interpretability. Rather than relying on \textit{post-hoc} saliency approximations, our architecture exposes multi-level decision pathways during forward aggregation: spatially biased self-attention captures lesion-level patch interactions, while modality-specific graph attention coefficients explicitly quantify the contribution of visual regions and clinical risk factors to each disease node. Evaluated on the BRSET dataset (16,266 photographs from 8,524 patients), the proposed model achieves $76.1\%$ Macro-AUPRC across seven conditions, matching or exceeding image-only ($75.9\%$) and ResNet-50+MLP fusion ($75.9\%$) baselines, with largest contextual gains in macular edema ($+3.5$ percentage points, pp) and diabetic retinopathy ($+2.1$ pp). Codebase is available at: \url{https://github.com/proshanto-c/ca-mlc-refactor}.
\end{abstract}


\section{Introduction}
\label{sec:introduction}

Fundus photography (FP) is widely used to screen for retinal conditions including increased cup-to-disc ratio (ICD), diabetic retinopathy (DR), and age-related macular degeneration (AMD)~\cite{TING2019100759}. Multiple pathologies can co-occur, motivating multi-label classification, with large-scale studies demonstrating the feasibility of joint retinal disease screening~\cite{cen2021automatic}. Beyond image appearance, patient history and systemic factors can provide complementary information, with multimodal approaches demonstrating the benefits of integrating clinical records into retinal disease prediction~\cite{10.1167/tvst.10.9.18}.

Automated retinal screening has traditionally used independent classifiers~\cite{gulshan2016development}, while recent multi-label approaches increasingly explore relationships between labels. LAGNet \cite{10635340} uses label-attention graphs to model disease correlations, while Label-Aware Dual graph neural networks (GNNs)~\cite{10670275} combine multimodal features with population-level and pathology graphs. Conventional multimodal approaches also combine pooled image features and tabular variables through feature fusion~\cite{10.1167/tvst.10.9.18}. However, such fusion baselines operate as opaque black boxes, relying on uncalibrated, \textit{post~hoc} saliency maps that cannot faithfully resolve fine-grained lesion interactions or explain cross-modal feature routing.

In this paper, we propose an examination-level heterogeneous GNN that decouples spatial fundus patches and clinical covariates into dedicated visual and clinical hubs with cross-modal message passing to seven disease nodes. A central contribution of our architecture is an attention-based design that provides intrinsic, multi-level interpretability without \textit{post hoc} attribution: spatially biased self-attention highlights fine-grained lesion-to-patch interactions, while Graph Attention Network (GAT) pooling explicitly reveals how spatial patches and clinical covariates are weighted during evidence aggregation. On BRSET~\cite{nakayama2023brset}, the model achieves $76.1\%$ Macro-Area Under Precision Recall Curve (AUPRC) across seven conditions, matching conventional fusion while surpassing image-only multi-label baselines in macular edema ($+3.5$ pp) and diabetic retinopathy ($+2.1$ pp).

\section{Methodology}
\label{sec:methodology}

\begin{figure*}[t]
    \centering
    \includegraphics[width=0.935\linewidth]{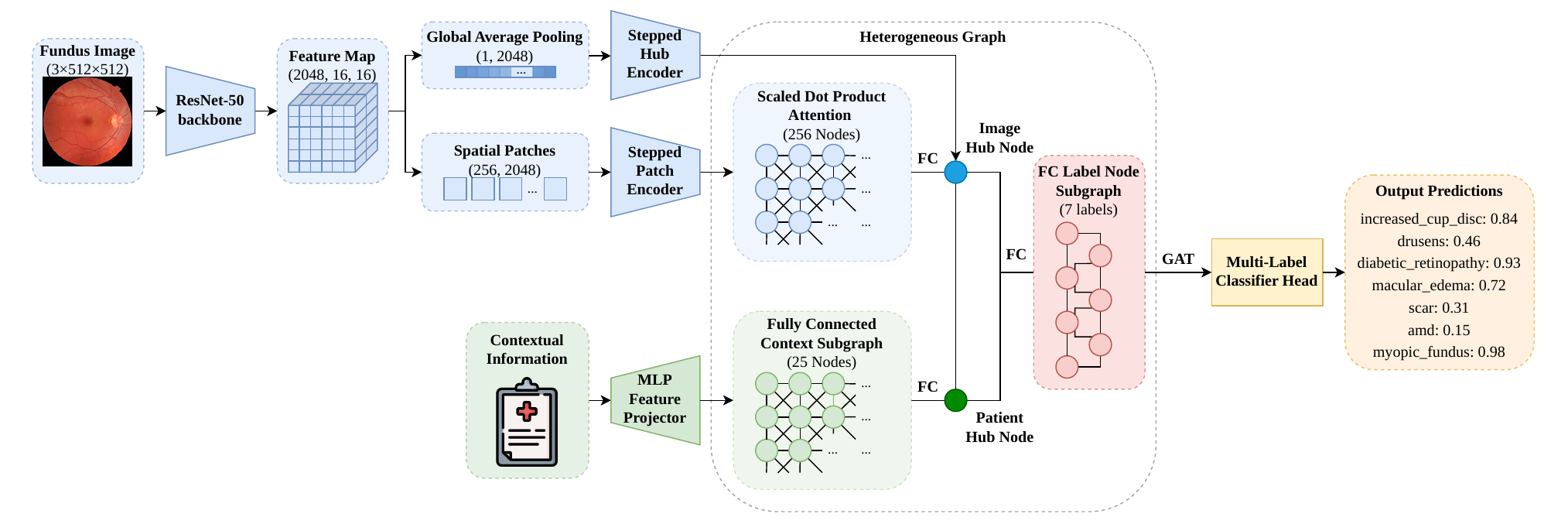}
    \vspace{-8pt}
    \caption{Architecture overview. Within each examination, patch tokens and clinical covariates are decoupled into modality hubs ($v_{\text{img}}$, $v_{\text{pat}}$) that exchange cross-modal evidence before routing to disease label nodes. FC edges denote full connectivity.}
    \label{fig:architecture}
    \vspace{-10pt}
\end{figure*}

We formulate multimodal retinal screening as an examination-level heterogeneous graph $\mathcal{G}=(\mathcal{V},\mathcal{E},\mathcal{R})$ comprising $|\mathcal{V}|=290$ nodes (256 spatial patches, 25 clinical covariates, 2 dedicated modality hubs $v_{\text{img}}$ and $v_{\text{pat}}$, and 7 disease label nodes; Fig.~\ref{fig:architecture}) that maintains separate visual and clinical representations before cross-modal communication.

\subsection{BRSET Dataset and Cohort Formulation}
\label{subsec:dataset}

We evaluate on the Brazilian Multilabel Ophthalmological Dataset (BRSET)~\cite{nakayama2023brset} ($16{,}266$ fundus photographs from $8{,}524$ patients). We evaluate 7 conditions with marked class imbalance: increased cup-to-disc ratio (ICD, $20.4\%$ test prevalence), drusen ($18.1\%$), diabetic retinopathy (DR, $6.8\%$), macular edema (ME, $2.6\%$), chorioretinal scar ($1.9\%$), age-related macular degeneration (AMD, $1.8\%$), and myopic fundus (MF, $1.5\%$).

Beyond retinal imagery, BRSET provides structured clinical history. We use 25 variables: age, sex, diabetes duration, insulin use, eye laterality, and binary indicators for the 20 most prevalent systemic comorbidities. In context-enabled graph models, these variables are represented as clinical nodes. Eye laterality is also embedded in the image hub. Image-only GNNs retain laterality in the image hub but omit other clinical variables; thus, ``Img'' denotes image features with examination laterality, not the complete absence of metadata. Patient-level multi-label stratification ($70\%/15\%/15\%$ train/validation/test) is used to ensure that all images pertaining to one patient are in the same partition.

\subsection{Spatially Biased Visual Encoding}
Given a fundus image $\mathbf{I}\in\mathbb{R}^{3\times512\times512}$, an ImageNet-pretrained ResNet-50 backbone (fine-tuned end-to-end) extracts feature map $\mathbf{F}\in\mathbb{R}^{2048\times16\times16}$, yielding $M=256$ patch tokens $\mathbf{f}_m$ and a global average-pooled descriptor $\mathbf{f}_{\text{glob}}$. Separate stepped multilayer perceptron (MLP) encoders project the patch tokens and global descriptor into a shared $d$-dimensional space, with $d$ selected from the tuning grid described in Sec.~\ref{sec:results_and_discussion}: $\mathbf{h}_{P,m}^{(0)} = \operatorname{MLP}_{\text{patch}}(\mathbf{f}_m)$ and $\mathbf{h}_{\text{img}}^{(0)} = \operatorname{MLP}_{\text{hub}}(\mathbf{f}_{\text{glob}}) + \mathbf{e}_{\text{eye}}$. 

Patch tokens undergo multi-head scaled dot-product attention (SDPA) with a 2D Gaussian spatial bias to model relationships between spatial locations:
\begin{equation}
    \mathbf{A}_{\text{intra}}^{(h)} = \operatorname{softmax}\!\left(
    \frac{\mathbf{Q}^{(h)}(\mathbf{K}^{(h)})^{\top}}{\sqrt{d_k}} + \mathbf{B}^{(h)}\right),
\end{equation}
where $B_{ij}^{(h)} = \gamma_h\exp(-\lVert\mathbf{p}_i-\mathbf{p}_j\rVert_2^2/\sigma_s^2)$, $\mathbf{p}_i$ is the 2D grid coordinate, $\sigma_s=8$, and $\gamma_h$ is a learnable head-specific scale initialized to 1.0. This adds a distance-dependent bias to the attention logits while retaining content-dependent interactions between spatially separated patches.

\subsection{Dual-Hub Heterogeneous GNN}
Continuous covariates (age, diabetes duration) are standardized using training statistics and paired with missingness flags $[z, m]$ before MLP projection; categorical variables are mapped through learned entity embeddings with an explicit unknown token. Disease label nodes are initialized as learned embedding vectors ($\mathbb{R}^d$) with target labels excluded from node inputs. Clinical and disease label nodes each form fully connected (FC) subgraphs with self-loops, weighted by signed Pearson correlations, $e_{ij}^{C}=\operatorname{Corr}(c_i,c_j)$ and $e_{ij}^{L}=\operatorname{Corr}(y_i,y_j)$, computed over the training set, retained as edge attributes.

At each of the $K=3$ heterogeneous layers, patch tokens are first updated via spatially biased self-attention (Eq.~1). Modality hubs ($v_{\text{img}}$ and $v_{\text{pat}}$) then aggregate evidence from their respective sub-graphs via GAT pooling, where scalar edge attributes $e_{uv}$ (unit weights for bipartite pooling, and signed Pearson correlations for clinical and label subgraphs) modulate attention logits to yield normalized aggregation coefficients $\alpha_{u \to v}$. A bidirectional image--patient bridge enables cross-modal information exchange between $v_{\text{img}}$ and $v_{\text{pat}}$, after which both hubs route information to the $L=7$ disease label nodes.

\subsection{Label-Specific Prediction}
Disease nodes produce multi-label predictions via $\hat{p}_{\ell} = \sigma(\mathbf{w}_{\ell}^{\top}\phi(\mathbf{h}_{\ell}^{(K)}) + b_{\ell})$, where $\phi(\cdot)$ is an MLP and output bias $b_{\ell}$ is initialized to empirical training log-odds $\log(N_\ell^+/(N - N_\ell^+))$ for $N_\ell^+$ positive cases out of $N$ training samples. We optimize positive-weighted binary cross-entropy ($\mathcal{L}_{\text{BCE}}$) with square-root dampened weights $w_\ell = \sqrt{\frac{N - N_\ell^+} {\max(N_\ell^+, 1)}}$ to address severe class imbalance. Per-condition decision thresholds $\tau_\ell^*$ are selected on the validation set to maximize $F_1$.

\section{Experiments and Results}
\label{sec:results_and_discussion}

\begin{table*}[t]
\caption{Test performance (mean $\pm$ std across five seeds, in \%). Context-Only, Single-Label (Img), Multi-Label (Img), and Dual-Hub are GNN configurations; ResNet-50 (Img+Ctx) is the ResNet-50+MLP feature fusion baseline. Bold denotes the best score among image-based and multimodal configurations.}
    \centering
    \resizebox{0.96\textwidth}{!}{%
    \setlength{\tabcolsep}{4.5pt}
    \begin{tabular}{l cc cc cc cc cc}
        \toprule
         & \multicolumn{2}{c}{\textbf{Context-Only}} & \multicolumn{2}{c}{\textbf{Single-Label (Img)}} & \multicolumn{2}{c}{\textbf{Multi-Label (Img)}} & \multicolumn{2}{c}{\textbf{ResNet-50 (Img+Ctx)}} & \multicolumn{2}{c}{\textbf{Dual-Hub GNN}} \\
        \cmidrule(lr){2-3} \cmidrule(lr){4-5} \cmidrule(lr){6-7} \cmidrule(lr){8-9} \cmidrule(lr){10-11}
        \textbf{Target} & \textbf{AUPRC} & \textbf{$F_1$} & \textbf{AUPRC} & \textbf{$F_1$} & \textbf{AUPRC} & \textbf{$F_1$} & \textbf{AUPRC} & \textbf{$F_1$} & \textbf{AUPRC} & \textbf{$F_1$} \\
        \midrule
        ICD           & $31.0 \pm 1.3$ & $39.8 \pm 0.9$ & $\mathbf{78.7} \pm 1.7$ & $\mathbf{71.2} \pm 1.4$ & $73.4 \pm 3.2$ & $67.9 \pm 3.3$ & $73.1 \pm 2.3$ & $68.2 \pm 1.7$ & $74.0 \pm 2.3$ & $68.6 \pm 1.5$ \\
        Drusen           & $33.5 \pm 0.5$ & $34.3 \pm 1.0$ & $\mathbf{76.4} \pm 0.7$ & $\mathbf{70.2} \pm 1.0$ & $75.2 \pm 1.8$ & $68.1 \pm 1.0$ & $72.2 \pm 2.3$ & $68.0 \pm 0.6$ & $75.3 \pm 1.4$ & $68.3 \pm 1.1$ \\
        DR            & $44.8 \pm 1.8$ & $51.0 \pm 2.2$ & $\mathbf{87.3} \pm 2.0$ & $\mathbf{81.3} \pm 3.3$ & $84.5 \pm 1.3$ & $78.8 \pm 2.3$ & $87.2 \pm 1.3$ & $81.2 \pm 1.0$ & $86.6 \pm 1.4$ & $81.2 \pm 1.5$ \\
        ME            & $27.2 \pm 4.8$ & $30.0 \pm 6.3$ & $77.1 \pm 5.7$ & $68.6 \pm 7.0$ & $81.7 \pm 2.4$ & $75.7 \pm 2.1$ & $\mathbf{86.0} \pm 2.1$ & $77.1 \pm 2.4$ & $85.2 \pm 0.8$ & $\mathbf{77.4} \pm 0.9$ \\
        Scar          & $2.9 \pm 0.5$ & $4.0 \pm 0.5$ & $83.7 \pm 1.9$ & $80.6 \pm 4.7$ & $\mathbf{84.2} \pm 3.4$ & $\mathbf{81.1} \pm 2.6$ & $83.0 \pm 4.3$ & $78.4 \pm 6.0$ & $83.8 \pm 1.7$ & $78.9 \pm 0.7$ \\
        AMD           & $14.9 \pm 0.7$ & $26.1 \pm 3.7$ & $56.7 \pm 2.7$ & $51.1 \pm 3.5$ & $55.3 \pm 6.5$ & $54.0 \pm 7.2$ & $\mathbf{57.4} \pm 3.4$ & $54.9 \pm 2.0$ & $53.4 \pm 7.1$ & $\mathbf{55.2} \pm 4.4$ \\
        MF            & $3.0 \pm 0.5$ & $4.5 \pm 2.0$ & $66.8 \pm 8.6$ & $61.9 \pm 8.3$ & $\mathbf{76.8} \pm 3.8$ & $\mathbf{69.1} \pm 5.9$ & $72.5 \pm 7.2$ & $67.4 \pm 7.0$ & $74.6 \pm 4.2$ & $64.6 \pm 6.9$ \\
        \midrule
        \textbf{Macro} & $22.5 \pm 0.7$ & $27.1 \pm 1.7$ & $75.2 \pm 1.3$ & $69.3 \pm 1.6$ & $75.9 \pm 0.6$ & $70.6 \pm 1.5$ & $75.9 \pm 1.5$ & $\mathbf{70.7} \pm 2.0$ & $\mathbf{76.1} \pm 1.1$ & $70.6 \pm 0.9$ \\
        \bottomrule
    \end{tabular}%
    }
    \label{tab:main_results}
\end{table*}


The Context-Only, Single-Label (Img), Multi-Label (Img), and Dual Hub configurations are GNNs implemented in PyTorch Geometric; the ResNet-50 (Img+Ctx) baseline is a ResNet-50+MLP model implemented in PyTorch. Models are trained across five random seeds for 150 epochs using positive-weighted binary cross-entropy and AdamW (learning rate $10^{-4}$, 15-epoch linear warmup). Images are resized to $512\times512$, horizontally flipped, and ImageNet-normalized. Checkpoints are selected by validation Macro-AUPRC. GNNs use three GAT layers ($K=3$, $H=2$ heads). Hyperparameters are selected by validation performance from $d\in\{128,256,512\}$ and dropout $p\in\{0.15,0.25,0.35\}$. For the independently trained single-label GNNs, hyperparameters were tuned separately for each target: MF and drusen selected $d=128$ and $p=0.25$, while the remaining targets selected $d=512$ and $p=0.15$. The multi-label GNNs use a shared configuration across their label nodes, selected at the model level.

The ResNet-50 (Img+Ctx) baseline concatenates global average-pooled ResNet features with a two-layer MLP projection of the clinical variables, providing a conventional feature fusion comparison. Decision thresholds $\tau_\ell^*$ are selected on validation $F_1$. We report AUPRC and $F_1$; AUROC exceeded $0.92$ across conditions and is omitted for brevity.

\subsection{Classification Performance}
\label{subsec:quantitative_results}

Table~\ref{tab:main_results} reports test performance for the GNN configurations and the ResNet-50+MLP baseline. The proposed dual-hub GNN obtains $76.1\pm1.1\%$ Macro-AUPRC, comparable to the ResNet-50+MLP feature fusion baseline ($75.9\pm1.5\%$). Conventional fusion leads on DR ($+0.6$ pp) and ME ($+0.8$ pp), while the GNN leads on four of the seven conditions, especially drusen ($+3.1$ pp) and MF ($+2.1$ pp). Neither approach consistently dominates, but the dual-hub GNN directly exposes patch-to-image-hub and clinical-to-patient-hub attention during aggregation, without requiring \textit{post hoc} attribution for these operations. This provides a structured alternative to conventional feature concatenation while maintaining competitive predictive performance.

The effect of adding clinical context varies by condition. Relative to the image-only multi-label GNN, context changes AUPRC by +0.6 pp (ICD), +0.1 (drusen), +2.1 (DR), +3.5 (ME), $-0.4$ (scar), $-1.9$ (AMD) and $-2.2$ (MF). Gains are largest for DR and ME, consistent with the potential relevance of diabetes-related covariates; contextual benefits vary across diseases.

\subsection{Attention Analysis}
\label{subsec:interpretability_results}

\begin{figure*}[t]
  \centering
  \includegraphics[width=0.82\linewidth]{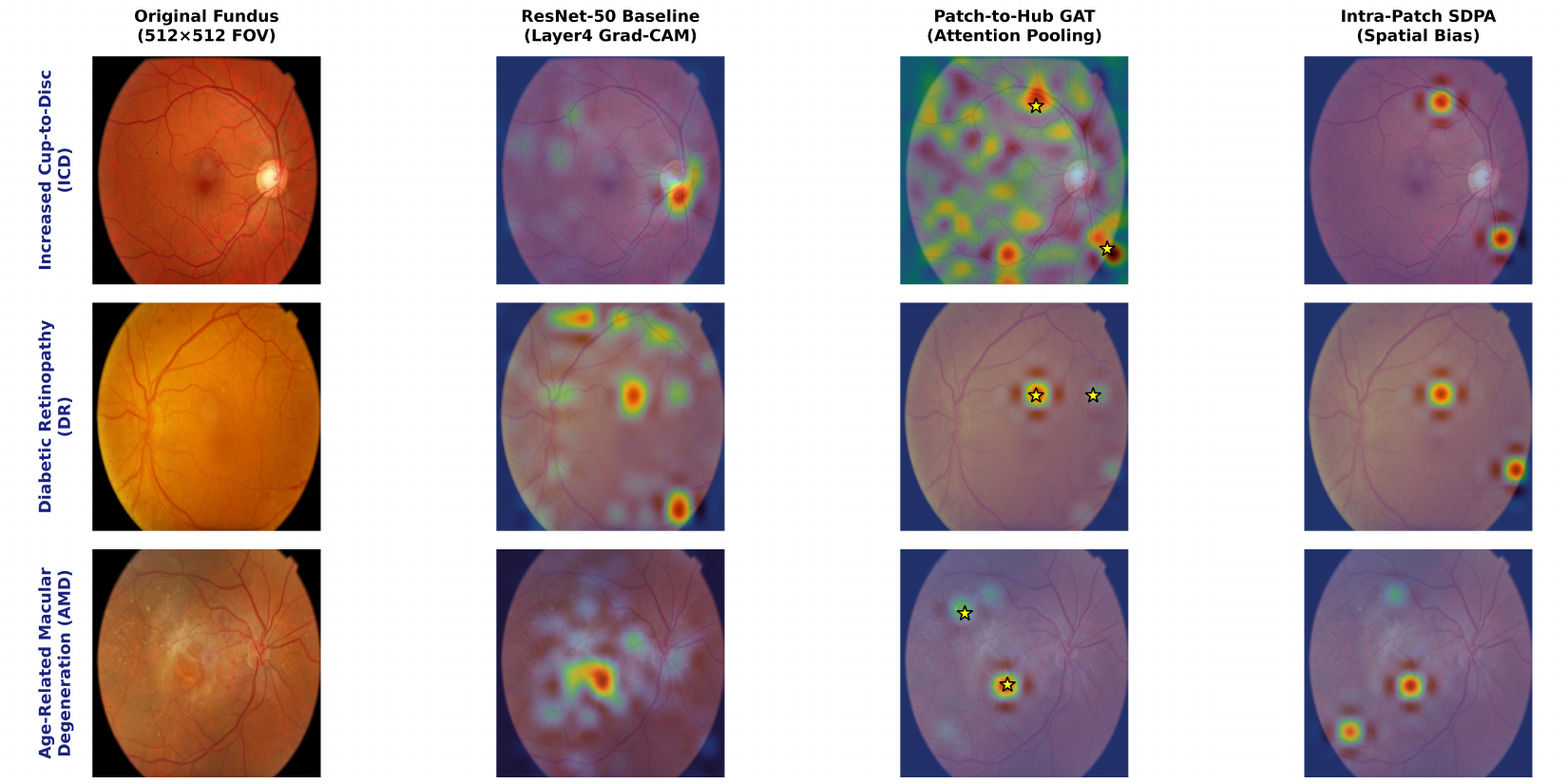}
  \vspace{-8pt}
  \caption{Visualisation of interpretability approaches on ICD, DR, and AMD examples. Original image and Grad-CAM on the ResNet50+MLP are Column 1 and 2 (left-to-right) respectively. Right two columns show Dual-Hub GNN attention visualisations. Column 3 shows attention for patch-to-hub edges. Column 4 shows patch-to-patch attention. Stars mark the two highest-attended GAT patches.}
  \label{fig:interpretability_comparison}
  \vspace{-10pt}
\end{figure*}

Figure~\ref{fig:interpretability_comparison} compares visual representations for representative ICD, DR, and AMD test images. Layer-0 patch-to-hub GAT attention ($\alpha_{p \to \text{img}}$) is averaged across heads, upsampled to $512\times512$ via bicubic spline interpolation with Gaussian smoothing ($\sigma=1.2$), and min-max normalized. Unlike \textit{post~hoc} Gradient-weight Class Activation Mapping (Grad-CAM) attribution, GAT attention directly represents the weighting of spatial features during patch-to-hub aggregation. In the examples shown, these attention maps concentrate on focal retinal regions, including the optic disc margin in ICD, micro-aneurysm and hemorrhage clusters in DR, and central drusen in AMD. Yellow stars indicate the two highest-attended patch locations. Spatially biased self-attention additionally visualizes interactions between patch tokens. The clinical branch similarly exposes attention coefficients ($\alpha_{c\to\text{pat}}$) over patient covariates, enabling direct inspection of modality-specific information aggregation.

\section{Discussion}
\label{subsec:discussion}

The results show that structured clinical context can be incorporated into multi-label retinal classification while maintaining performance comparable to image-only and conventional feature fusion approaches. The dual-hub design preserves separate spatial and clinical representations while enabling information exchange before disease prediction. This provides a structured alternative to direct feature concatenation and exposes modality-specific attention during aggregation. The disease-dependent performance changes further suggest that the value of clinical context depends on the relationship between recorded covariates and target pathology.

\section{Conclusion}
\label{sec:conclusion}

We present a dual-hub heterogeneous GNN for context-aware multi-label retinal classification. The model achieves $76.1\%$ Macro-AUPRC, comparable to the image-only multi-label GNN and ResNet-50+MLP feature fusion baseline, with the largest gains over the image-only multi-label model observed for DR and ME. By separating fundus images and clinical history into dedicated, interacting hubs, our model predicts co-occurring diseases while keeping the cross-modal attention weights fully interpretable.

\section{Compliance with Ethical Standards}
This study used retrospective, de-identified BRSET data accessed under the PhysioNet Credentialed Health Data License. Ethical approval was not required, as confirmed by the license attached to the data.

\section{Acknowledgments}
The authors acknowledge the use of the services/facilities of the Institute of Biomedical Engineering (IBME), Department of Engineering Science, University of Oxford. The authors express no conflicts of interest. PC is supported by the Crankstart Graduate Scholarship in partnership with the EPSRC Doctoral Landscape Award. TZ is supported by the Royal Academy of Engineering under the Research Fellowship scheme. AB is supported by the Royal Society University Research Fellowship (Grant No. URF\textbackslash R1\textbackslash 221314).

\bibliographystyle{IEEEbib}
\bibliography{refs}

@article{TING2019100759,
title = {Deep learning in ophthalmology: The technical and clinical considerations},
journal = {Progress in Retinal and Eye Research},
volume = {72},
pages = {100759},
year = {2019},
issn = {1350-9462},
doi = {https://doi.org/10.1016/j.preteyeres.2019.04.003},
url = {https://www.sciencedirect.com/science/article/pii/S1350946218300909},
author = {Ting, D.S.W. and others}
}

@article{10.1167/tvst.10.9.18,
    author = {Hsu, M. and others},
    title = {Deep Learning for Automated Diabetic Retinopathy Screening Fused With Heterogeneous Data From {EHRs} Can Lead to Earlier Referral Decisions},
    journal = {Translational Vision Science \& Technology},
    volume = {10},
    number = {9},
    pages = {18-18},
    year = {2021},
    month = {08},
    issn = {2164-2591},
    doi = {10.1167/tvst.10.9.18},
    url = {https://doi.org/10.1167/tvst.10.9.18},
    eprint = {https://arvojournals.org/arvo/content_public/journal/tvst/938550/i2164-2591-10-9-18_1629788375.00135.pdf},
}

@INPROCEEDINGS{10635340,
  author={Liang, H. and others},
  booktitle={2024 IEEE International Symposium on Biomedical Imaging (ISBI)}, 
  title={{LAGNet}: Label Attention Graph Networks for Ocular Disease Classification Using Fundus Images}, 
  year={2024},
  volume={},
  number={},
  pages={1-5},
  doi={10.1109/ISBI56570.2024.10635340}}

@ARTICLE{10670275,
  author={Liu, Y. and others},
  journal={IEEE Journal of Biomedical and Health Informatics}, 
  title={Label-Aware Dual Graph Neural Networks for Multi-Label Fundus Image Classification}, 
  year={2025},
  volume={29},
  number={4},
  pages={2731-2743},
  doi={10.1109/JBHI.2024.3457232}}

@article{nakayama2023brset,
  author    = {Nakayama, L.F. and others},
  title     = {{BRSET}: A {Brazilian} Multilabel Ophthalmological Dataset of Retina Fundus Photos},
  journal   = {PLOS Digital Health},
  volume    = {3},
  number    = {7},
  pages     = {e0000454},
  year      = {2024},
  doi       = {10.1371/journal.pdig.0000454}
}

@article{cen2021automatic,
  author  = {Cen, L.-P. and others},
  title   = {Automatic Detection of 39 Fundus Diseases
             and Conditions in Retinal Photographs
             Using Deep Neural Networks},
  journal = {Nature Communications},
  volume  = {12},
  pages   = {4828},
  year    = {2021},
  doi     = {10.1038/s41467-021-25138-w}
}

@article{gulshan2016development,
  author  = {Gulshan, V. and others},
  title   = {Development and Validation of a Deep
             Learning Algorithm for Detection of
             Diabetic Retinopathy in Retinal Fundus
             Photographs},
  journal = {JAMA},
  volume  = {316},
  number  = {22},
  pages   = {2402--2410},
  year    = {2016},
  doi     = {10.1001/jama.2016.17216}
}

\end{document}